\documentclass[11pt, aps,english,nofootinbib, showkeys,a4]{revtex4-1}

\usepackage[T1]{fontenc}
\usepackage[utf8]{inputenc}
\usepackage{xcolor}

\usepackage{babel}
\usepackage{pmboxdraw}
\usepackage{amstext}
\usepackage{graphicx}
\usepackage{bm}
\usepackage{amsmath}
\usepackage{makecell}
\usepackage{booktabs}
\usepackage[version=4]{mhchem}
\usepackage{float}
\usepackage{chemformula} 
\usepackage[T1]{fontenc} 
\usepackage{booktabs} 
\usepackage{makecell} 
\usepackage{xcolor}
\usepackage{comment}
\usepackage{booktabs}
\usepackage{array}
\usepackage{tabularx}
\usepackage{array} 
\usepackage{makecell} 
\usepackage{hyperref}
\usepackage[official]{eurosym}
\usepackage{rotating}
\usepackage{placeins}
\usepackage{multirow}
\usepackage{subcaption}

\begin{document}

\title{Machine Learning-Based Manufacturing Cost Prediction from 2D Engineering Drawings via Geometric Features}

\author{Ahmet Bilal Ar{\i}kan}
\affiliation{Institute for Data Science \& Artificial Intelligence, Bo\u{g}azi\c{c}i University,
\.{I}stanbul, T\"{u}rkiye}

\author{\c{S}ener \"{O}z\"{o}nder}
\email{Corresponding author: sener.ozonder@bogazici.edu.tr}
\affiliation{Institute for Data Science \& Artificial Intelligence, Bo\u{g}azi\c{c}i University,
\.{I}stanbul, T\"{u}rkiye}
\email{Corresponding author: sener.ozonder@bogazici.edu.tr}

\author{H. K\"{u}bra K\"{u}\c{c}\"{u}kkartal}
\affiliation{ArtificaX Technologies, Bo\u{g}azi\c{c}i Teknopark, \.{I}stanbul, T\"{u}rkiye}

\author{Murat Arslano\u{g}lu}
\affiliation{Teknorot Automative Products Inc., D\"{u}zce, T\"{u}rkiye}

\author{Fatih \c{C}a\u{g}{\i}rankaya}
\affiliation{Teknorot Automative Products Inc., D\"{u}zce, T\"{u}rkiye}

\author{Berk Ayvaz}
\affiliation{\.{I}stanbul Ticaret University, Faculty of Engineering, Industrial Engineering,
\.{I}stanbul, T\"{u}rkiye}

\author{Mustafa Taha Ko\c{c}yi\u{g}it}
\affiliation{Institute for Data Science \& Artificial Intelligence, Bo\u{g}azi\c{c}i University,
\.{I}stanbul, T\"{u}rkiye}

\author{H\"{u}seyin Oktay Altun}
\affiliation{Institute for Data Science \& Artificial Intelligence, Bo\u{g}azi\c{c}i University,
\.{I}stanbul, T\"{u}rkiye}

\begin{abstract}

We present an integrated machine learning framework that transforms how manufacturing cost is estimated from 2D engineering drawings. Unlike traditional quotation workflows that require labor-intensive process planning, our approach about 200 geometric and statistical descriptors directly from 13,684 DWG drawings of automotive suspension and steering parts spanning 24 product groups. Gradient-boosted decision tree models (XGBoost, CatBoost, LightGBM) trained on these features achieve nearly 10\% mean absolute percentage error across groups, demonstrating robust scalability beyond part-specific heuristics. By coupling cost prediction with explainability tools such as SHAP, the framework identifies geometric design drivers including rotated dimension maxima, arc statistics and divergence metrics, offering actionable insights for cost-aware design. This end-to-end CAD-to-cost pipeline shortens quotation lead times, ensures consistent and transparent cost assessments across part families and provides a deployable pathway toward real-time, ERP-integrated decision support in Industry 4.0 manufacturing environments.

\end{abstract}

\maketitle

\section{Introduction}

In manufacturing, precise estimation of production costs directly influences profitability, competitiveness and strategic decision-making \citep{WOS:001267487000001,Rounaghi2021}. Two-dimensional (2D) engineering drawings, particularly in DWG format, are foundational documents in the manufacturing pipeline, extensively used across various industries, including automotive, aerospace, electronics and heavy machinery \citep{10.1007/978-3-031-18326-3_36}. These drawings contain critical geometrical information, such as dimensions, tolerances and material specifications, serving as the initial step for production planning, machining operations, and cost estimation \citep{Lin2023}.

Cost estimation involves assessing both direct and indirect production costs. Direct costs encompass labor, raw materials and direct machine usage, while indirect costs cover overhead, maintenance, utilities and depreciation \citep{Li2005_CostEstimation}. Traditional cost estimation methods vary from simple empirical calculations to detailed analytical estimations, often relying on expert judgment or historical data \citep{Niazi}. Typically, the direct machining cost calculation involves determining the appropriate shop rate (hourly cost combining labor, machinery and indirect expenses) and multiplying it by the machining time \citep{Armillotta2021}.

However, accurate cost estimation for precision machined components presents significant challenges. Precisely forecasting machining time requires detailed process planning, careful selection of cutting parameters (e.g., cutting speed, feed rate, depth of cut) for various operations such as turning, drilling, milling or grinding, and accounting for non-productive handling and setup times \citep{Armillotta2021}. These complexities frequently result in inconsistent estimations, human errors and prolonged quotation periods. Furthermore, traditional linear regression models often fail to account for complex, nontrivial correlations and inherent nonlinearities embedded in production data, thereby diminishing their predictive accuracy \citep{Shamim}.

Given these limitations, sophisticated machine learning (ML) methods such as regression tree-based models—particularly XGBoost regressors—emerge as promising solutions. By leveraging historical sales prices, where both direct and indirect costs are inherently embedded, these ML models effectively capture complex relationships within the data, eliminating the need for heuristic or rule-based approaches, which are often complex and intractable. Periodic retraining ensures the model maintains high predictive accuracy and dynamically adapts to shifts in market conditions and production processes \citep{ijtech-5675,Shamim}.

Manual cost estimation, typically requiring teams of specialists dedicating several weeks per drawing, substantially elevates both quotation times and operational costs. This labor-intensive approach not only reduces overall competitiveness but also contributes to significant inconsistency and variability in estimated costs due to human errors.

The industrial demand is rapidly moving toward fast, accurate and fully automated cost estimation, consistent with Industry 4.0 principles and cybermanufacturing environments \citep{KHAN2025149}. A fully automated approach that relies solely on technical drawings—without the need for detailed production routes or supplementary process information—can directly address this requirement, improving efficiency and responsiveness, particularly in on-demand manufacturing scenarios \citep{Kusiak2018SmartM,Lasi2014}.

In the highly competitive automotive supply sector, where companies serve both domestic and international OEMs, speed in generating quotations is a decisive factor for winning contracts. However, product diversity and geometric complexity often make it difficult to meet the industry target of delivering quotations within 48 hours, with many estimates taking over a week. The manufacturing-on-demand business model, driven by digitalization, increasingly demands near-instant cost feedback. In this context, AI-driven quotation systems capable of generating reliable cost estimates within seconds can provide a substantial competitive advantage, enabling manufacturers to secure orders more quickly while optimizing design and procurement decisions.

This study proposes a novel machine learning-driven methodology for estimating production costs directly from 2D DWG engineering drawings. DWG files are systematically parsed to extract geometric features including line lengths, circle radii, angles, arc dimensions and material information. This feature extraction creates an extensive dataset (approximately 200 features per product group), subsequently employed to train gradient boosting decision tree models.

Our approach offers multiple advantages. Firstly, decision tree models, such as XGBoost, provide inherent explainability, facilitating insights into feature importance. Designers can thus identify and avoid specific geometric attributes that disproportionately increase manufacturing costs, fostering more cost-efficient design decisions. Secondly, the model significantly accelerates the quotation process for online manufacturing platforms, enabling immediate, accurate cost predictions from uploaded engineering drawings, essential for real-time customer feedback and streamlined operations \citep{WU20151}. Thirdly, our method enhances competitiveness by reducing quotation lead time and ensuring cost consistency across diverse product lines. Finally, as an indirect benefit, discrepancies highlighted by high prediction errors can signal inaccuracies in past manual estimations, prompting companies toward strategic reassessment and digital transformation initiatives.

\section{Related Works}

While 3D CAD models are increasingly common, 2D engineering drawings remain a staple in many manufacturing workflows (e.g. for CNC machining jobs). However, the academic literature on automated cost estimation directly from 2D drawings is relatively sparse, with most AI-based cost estimation studies focusing on 3D CAD models \citep{Lee2022,Zhang2018,Ning2020,Yoo_2021,CHAN2018115}. Traditional practice often involves experts manually interpreting 2D drawings to identify part features, dimensions and tolerances, then estimating costs based on experience or lookup tables \citep{SAJADFAR2015633}. This manual approach can be time-consuming and inconsistent \citep{LIU20138,Cavalieri2004}. A few research efforts have tackled the challenge of extracting cost-driving features from 2D drawings \citep{VanDaele_Decleyre_Dubois_Meert_2021,10.1007/978-3-031-18326-3_36,LIU20101907}
. Serrat et al. (2013) provide a notable example, focusing on custom industrial hoses. They developed a method to derive a 3D representation of a hose either from an STL file or from the orthographic projections in a 2D CAD drawing, and then estimate manufacturing time and cost from that geometry \citep{9dfd9c4eddff428bbd99a595b7eddc44}. Other studies have adopted graph- and ontology-based approaches to extract cost-related information from 2D drawings \citep{XIE2022103697,zhang2023component,KASIMOV2015134,WANG20141041}.

Research in additive manufacturing has predominantly addressed build duration rather than direct cost estimation, with various techniques—particularly analytical models that leverage process variables like laser power, scan speed, toolpath layout and layer thickness—developed to predict how long a build will take \citep{doi:10.1177/0954405416640661,Nishida_2019ijat,Gadelmawla2013,Gurgenc2019,HEO2006437,LIU20138}. Purely analytical machining‐time predictions prove impractical because obtaining the extensive data required—machine dynamics (acceleration/deceleration profiles, structural limits), CNC control specifics (block processing times, jerk limitations, sampling intervals) and the precise configuration of the controller within the machine tool (movement accuracy, positional precision, and related settings)—is exceedingly difficult \citep{Coelho2010-sx}.

Part designers and cost engineers are often interested in which sections of the drawings control the cost the most. One straightforward way to maintain interpretability is to use models that are inherently transparent, such as decision trees or rule-based systems. Decision tree learning algorithms create human-readable if-then rules (splitting on feature thresholds) that make their reasoning easy to follow \citep{mourikas2017machine}. In the early 2000s, some cost estimation systems were built as expert systems or decision trees. For example, an expert system might encode rules like ``IF material = steel AND number of holes $>$ 10 AND tolerance $<$ 0.1 mm, THEN add 20\% to base cost.'' While such rules can be effective for well-understood domains, they are labor-intensive to create and maintain. Nonetheless, decision tree-based models have been noted for their clarity in cost modeling contexts \citep{mourikas2017machine}.
Even when more complex ML models are used, practitioners sometimes extract a simplified decision tree from them for explanation. In manufacturing cost estimation, these interpretable models have seen use in applications such as process selection and supplier quotation, where justifying the estimate to customers is important.

Beyond plain decision trees, researchers have explored ensemble methods that retain some interpretability. Gradient boosting machine implementations like XGBoost allow analysis of feature importance scores, partial dependence plots and other techniques to interpret the model's behavior. Mandolini et al. (2024) recently emphasized the importance of making ML cost models explainable to engineers \citep{WOS:001267487000001}.  They presented a cost modeling framework for engineered-to-order products that uses an automated cost estimation tool to generate training data (i.e., simulating various design scenarios) and then trains ML models. Importantly, they report that their resulting cost models are “explainable (i.e., not conceived as a black box)” and provide engineers with a ranked list of cost-driving features for a design \citep{Yoo_2021}.
This was achieved by using algorithms like random forest or gradient boosting and focusing on feature selection and importance – essentially ensuring the model could point to which input features (dimensions, material, etc.) were raising the cost in each case. The benefit is that design teams can use this information to modify designs to reduce cost. For example, if the model shows that a small fillet radius is causing a cost spike due to a difficult machining operation, the designer might increase that radius, if possible \citep{WOS:001267487000001}.

\section{Dataset}

The data preprocessing pipeline for converting raw engineering drawings into a structured tabular dataset includes the collection and preprocessing of DWG files, extraction of geometric features from 2D drawings and construction of both descriptive and statistical representations. The processed dataset includes shape-based features, material encodings and computed distance metrics. The target variable is the manufacturing cost, which is derived from the historical production data updated for inflation to present‑day price levels.

\subsection{Dataset Overview}

\begin{figure}[htbp]
\centering
\includegraphics[width=\linewidth]{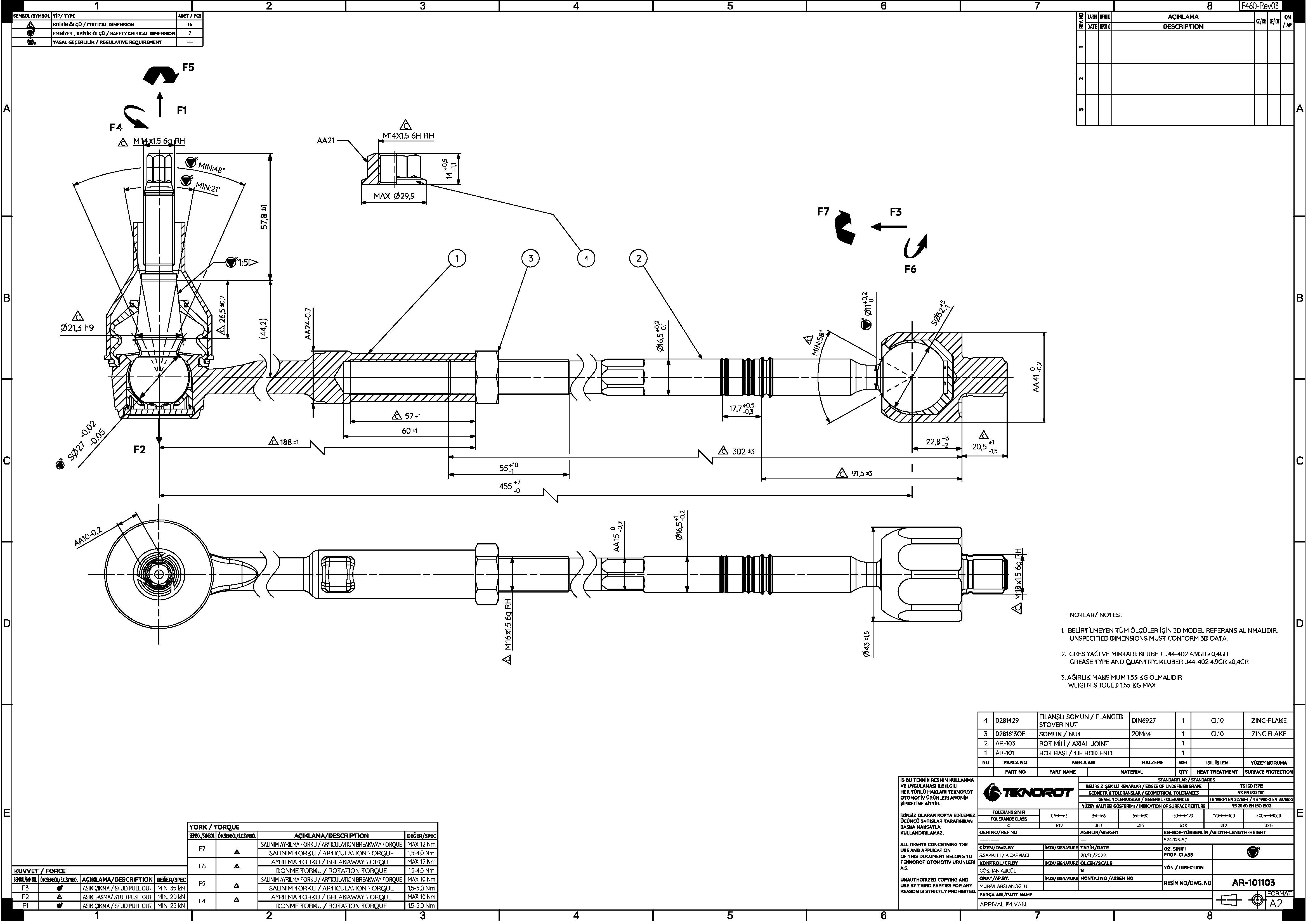}
\caption{Example engineering drawing from the dataset.}
\label{drawing}
\end{figure}

The dataset consists of 13,684 2D engineering drawings provided in the DWG file format, 
representing automotive suspension and steering components from 30 different car makers. These drawings cover a wide range of mechanical parts organized into 24 distinct product groups. Each DWG file corresponds to a component or complete assembly and is associated with a historical unit manufacturing cost recorded in Euros, ranging from €0.50 to €50.00. An example engineering drawing from the dataset is shown in Figure~\ref{drawing}, illustrating the typical structure and annotation style present in the raw DWG files.

\FloatBarrier

\FloatBarrier

To facilitate parsing, all DWG files were first batch-converted to the DXF format, which is a raw text format. This conversion step was essential to ensure reliable and standardized extraction of geometric entities. This approach has superiority to vision based methods because it preserves exact geometric and semantic information from the drawing, avoiding potential inaccuracies introduced by image rendering, resolution limits or computer vision feature detection.

\begin{figure*}[htbp]
  \centering
  \begin{subfigure}[b]{0.30\textwidth}
    \includegraphics[width=\linewidth]{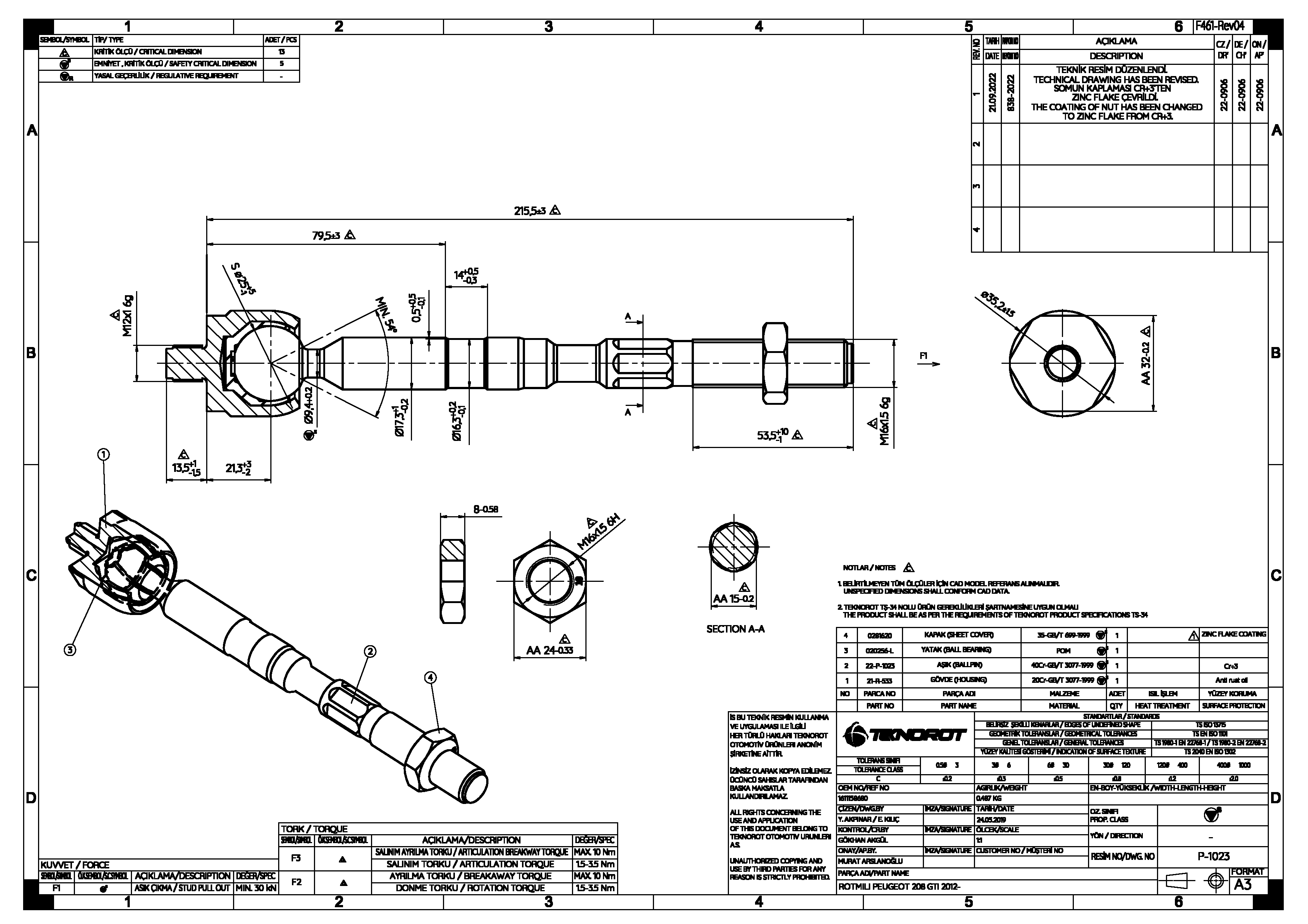}
    \caption{Inner Tie Rod (960)}
  \end{subfigure}\hfill
  \begin{subfigure}[b]{0.30\textwidth}
    \includegraphics[width=\linewidth]{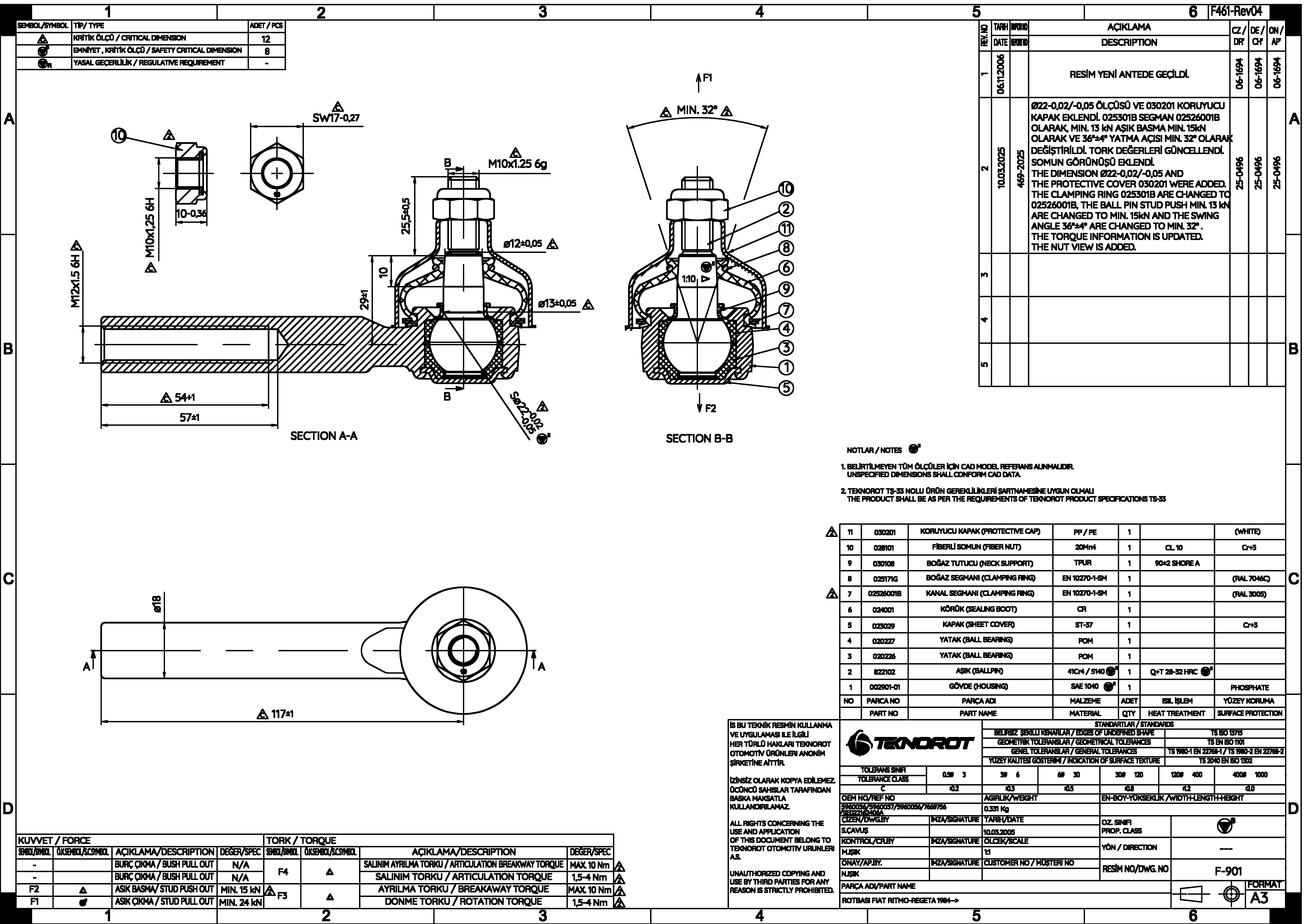}
    \caption{Tie Rod End (1157)}
  \end{subfigure}\hfill
  \begin{subfigure}[b]{0.30\textwidth}
    \includegraphics[width=\linewidth]{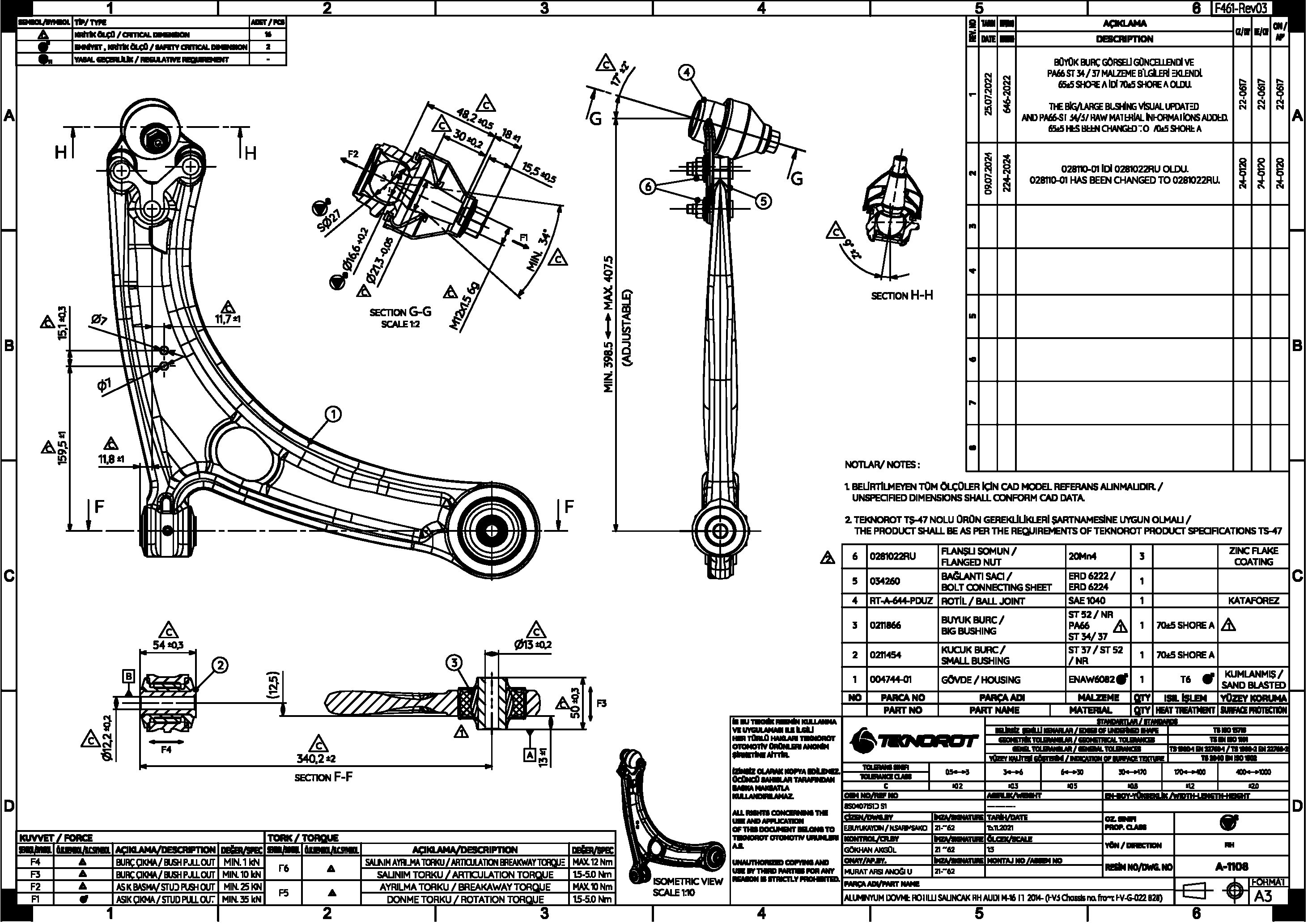}
    \caption{Control Arms (651)}
  \end{subfigure}

  \vspace{1ex}

  \begin{subfigure}[b]{0.30\textwidth}
    \includegraphics[width=\linewidth]{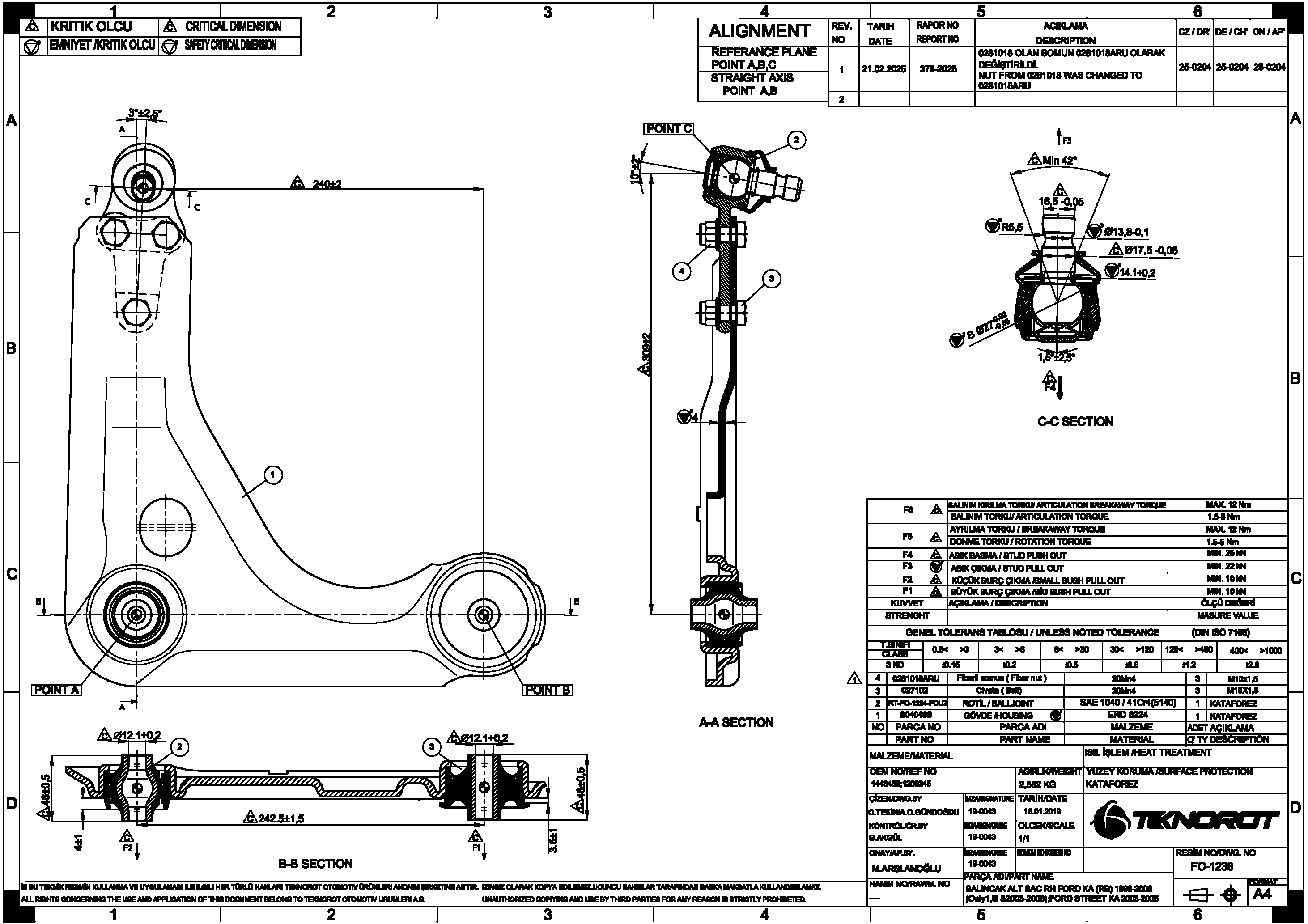}
    \caption{Wishbones (1604)}
  \end{subfigure}\hfill
  \begin{subfigure}[b]{0.30\textwidth}
    \includegraphics[width=\linewidth]{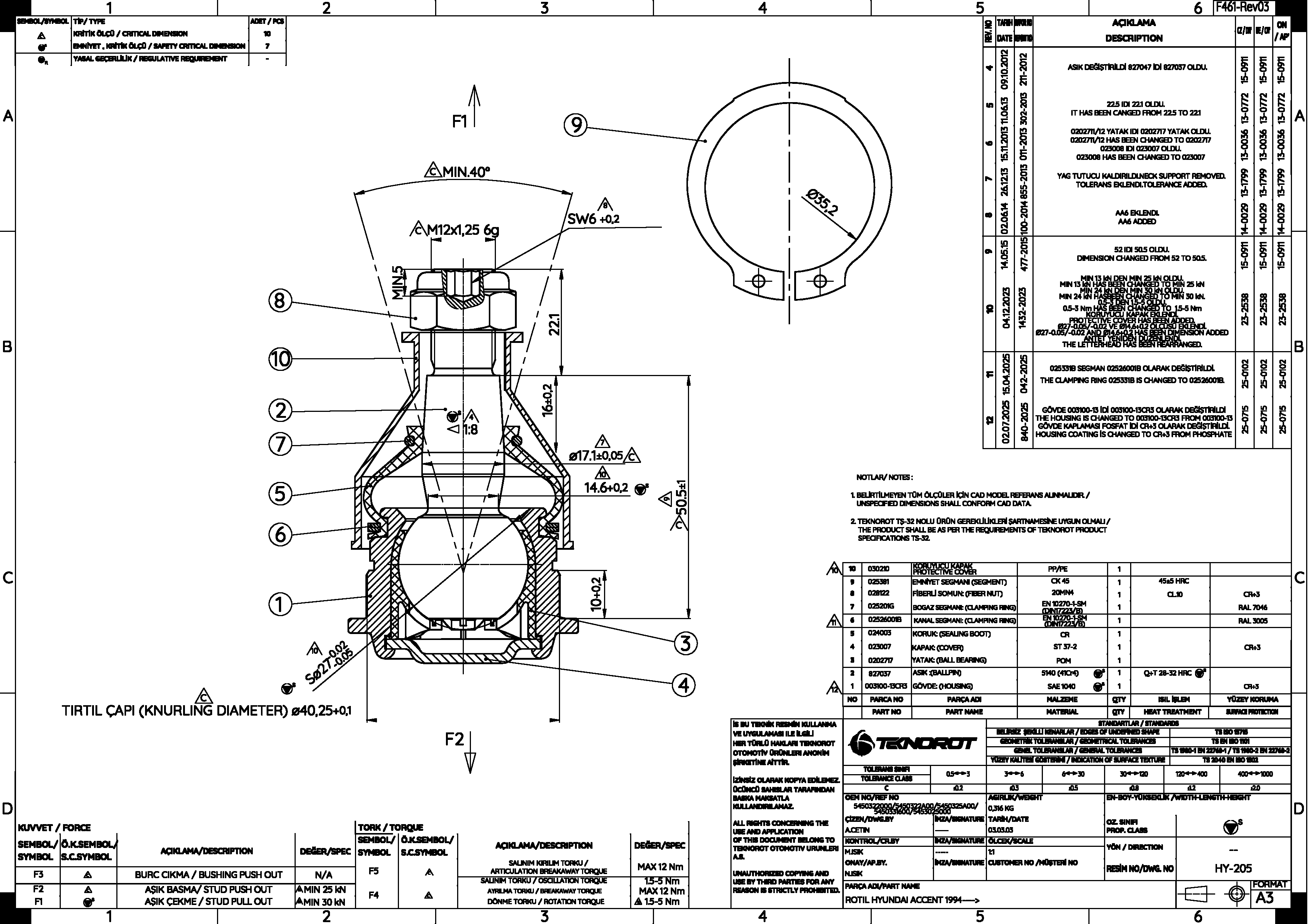}
    \caption{Ball Joint (811)}
  \end{subfigure}\hfill
  \begin{subfigure}[b]{0.30\textwidth}
    \includegraphics[width=\linewidth]{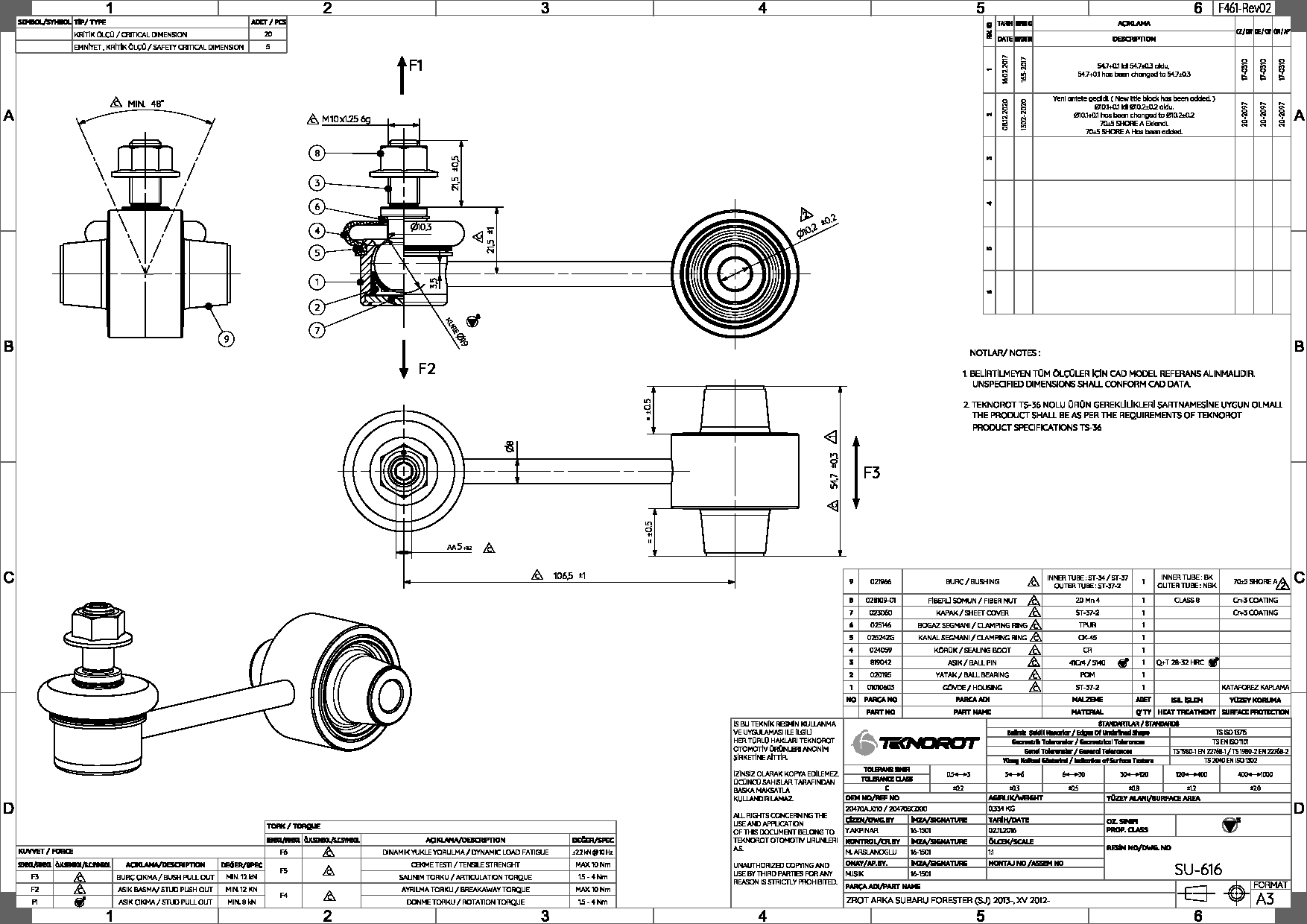}
    \caption{Link Stabilizer (1553)}
  \end{subfigure}

  \caption{Representative examples of drawings from different product groups along with number of drawings in each group.}
  \label{fig:product_summary}
\end{figure*}

The product groups vary in geometric complexity, material composition and manufacturing cost. Figure~\ref{fig:product_summary} presents a visual overview of selected product categories from the dataset, with each representative drawing accompanied by the corresponding product group name and the number of available DWG files.

\FloatBarrier
\FloatBarrier

\subsection{Geometrical Parsing}

To extract meaningful features from raw engineering drawings, we developed a geometrical parsing pipeline. The parsing process focused on three main categories of entities:

(1) Geometric shapes: We parsed five primary geometric primitives—\texttt{LINE}, \texttt{CIRCLE}, \texttt{ARC}, \texttt{SPLINE}, and \texttt{ELLIPSE}. For each shape type, relevant geometric properties were extracted, such as start and end points, center coordinates and radii where applicable. This ensured that the geometric identity and dimensional information of each element were fully captured. A summary of the parsed properties for each entity is provided in Table~\ref{tab:parsed_properties}.

\FloatBarrier

\begin{table}[htbp]
\centering
\caption{Parsed properties for each geometric entity.}
\renewcommand\arraystretch{1.3}
\begin{tabular}{@{}ll@{}}
\toprule
\textbf{Entity Type} & \textbf{Properties} \\ 
\midrule
Line    & Start point, end point, length \\ 
Circle  & Center point, radius \\ 
Arc     & Center point, radius, start angle, end angle \\ 
Spline  & Control points, fit points, degree \\ 
Ellipse & Center point, major axis, minor axis \\ 
\bottomrule
\end{tabular}

\label{tab:parsed_properties}
\end{table}

\FloatBarrier

(2) Dimensions: We extracted dimension entities such as Rotated, Angular, Diametric and Radial Dimensions, which convey measurement information within the drawings. For each, we parsed the dimension type, displayed measurement text, actual measurement and associated tolerances. Additionally, a global scale factor was computed using the measurement from the Rotated Dimension. This scale was then used to convert the lengths and sizes of geometric entities into real-world units.

(3) Textual content: Material texts embedded in \texttt{TEXT} and \texttt{MTEXT} entities were extracted. These material labels serve as categorical features for modeling.

 The output of this stage is a set of structured features for each drawing, encompassing geometric entities, dimensional measurements and material information. These serve as the foundation for the feature engineering process.

\subsection{Feature Engineering}

The first stage of feature engineering involved computing descriptive and statistical features from the parsed geometric and dimensional entities. These features capture the distribution and complexity of shapes within each drawing and are critical for  cost estimation. Summary statistics were computed specifically for the following geometric and dimensional quantities: line lengths, arc lengths, arc angles (i.e., total angular span per arc), circle radii and dimension measurements. All geometric measurements related to length and size were scaled using the computed factor to ensure consistency in physical units across drawings. For each of these quantities, we extracted a set of statistical descriptors, including
count, min, max, range, mean, median, mode, standard deviation, skewness and kurtosis. To capture the distributional characteristics in more detail, we additionally computed histogram counts (12 bins) and normalized histogram proportions (12 bins). These features provide a robust characterization of the shape and scale of components within each drawing and are foundational for representing part complexity and size variation in the modeling pipeline.

In the second phase of feature engineering, we computed distribution-based distance metrics to quantify how each drawing deviates from the typical shape characteristics of its corresponding product group. Then, for each product group, we calculated the average bin values across all drawings to represent the reference distribution for that group and shape. Using these group-level reference distributions, we computed the Euclidean distance and Kullback–Leibler (KL) divergence between each individual drawing's histogram and the corresponding group mean. The formulas for computing the Euclidean distance and KL divergence between these distributions are provided in Table~\ref{tab:distance_metrics}. Here \(\text{bins}_{\text{dwg},i}\) denote the normalized frequency in bin \(i\) for a given drawing, and \(\text{bins}_{\text{mean},i}\) denote the mean frequency in bin \(i\) for its corresponding product group. These distance-based features help capture how typical or anomalous a given drawing is within its product family in terms of geometric proportions.

\begin{table}[htbp]
\centering
\caption{Mathematical definitions of the distance metrics computed between each drawing's histogram bins and the average distribution of its product group.}
\label{tab:distance_metrics}
\begin{tabular}{@{}l l@{}}
\toprule
\textbf{Metric} & \textbf{Formula} \\ 
\midrule
Euclidean distance & 
\(
D_{\mathrm{Euc}}(\mathrm{bins}_{\mathrm{dwg}}, \mathrm{bins}_{\mathrm{mean}})
= \sqrt{\sum_{i=1}^{12}\bigl(\mathrm{bins}_{\mathrm{dwg},i} - \mathrm{bins}_{\mathrm{mean},i}\bigr)^{2}}
\) \\[1ex]
KL divergence & 
\(
D_{\mathrm{KL}}(\mathrm{bins}_{\mathrm{dwg}}, \mathrm{bins}_{\mathrm{mean}})
= \sum_{i=1}^{12} \mathrm{bins}_{\mathrm{mean},i}
\log\!\bigl(\tfrac{\mathrm{bins}_{\mathrm{mean},i} + \epsilon}{\mathrm{bins}_{\mathrm{dwg},i} + \epsilon}\bigr)
\) \\
\bottomrule
\end{tabular}
\end{table}

\section{Machine Learning Approach and Implementation}

We employed three state-of-the-art gradient boosting algorithms \citep{10.3389/fnbot.2013.00021} for regression: XGBoost \citep{10.1145/2939672.2939785}, CatBoost \citep{NEURIPS2018_14491b75} and LightGBM \citep{NIPS2017_6449f44a}. These models were selected due to their strong performance on tabular data \citep{SHWARTZZIV202284}, robustness to multicollinearity and built-in handling capability of missing values \citep{Pokhrel_Lazar_2022}. Specifically, gradient boosting based decision tree methods naturally capture nonlinear relationships and interaction effects \citep{WOS:000173361700001}, and they are resilient to multicollinearity—eliminating the need for explicit feature decorrelation.

All models were trained to predict the unit manufacturing cost using the features extracted from DXF files. Train, validation and test sets were split in a 70:15:15 ratio. Hyperparameter optimization was conducted separately for each product type using Bayesian optimization via Optuna \citep{akiba2019optunanextgenerationhyperparameteroptimization}, with the mean absolute percentage error (MAPE) as the optimization objective. A 5-fold cross-validation strategy was employed to ensure robust evaluation and reduce the risk of overfitting. Early stopping with a patience of 20 rounds was applied during training to prevent unnecessary computation.

Each model utilized its respective boosting mechanism and loss function: XGBoost and LightGBM used mean squared error (MSE) loss, while CatBoost used root mean squared error (RMSE). Hyperparameters such as learning rate, tree depth and regularization strengths were tuned, along with model-specific parameters like \texttt{bagging\_temperature} and \texttt{random\_strength} for CatBoost, and \texttt{num\_leaves} and \texttt{min\_child\_samples} for LightGBM. A summary of the tuned hyperparameters is provided in Table~\ref{tab:hyperparam_tuning}.

\FloatBarrier
\begin{table}[htbp]
\centering
\caption{Ranges and settings of hyperparameters tuned for each gradient-boosting algorithm.}
\label{tab:hyperparam_tuning}
\begin{tabular}{@{}l c c c@{}}
\toprule
\textbf{Hyperparameter} & \textbf{XGBoost} & \textbf{CatBoost} & \textbf{LightGBM} \\
\midrule
\texttt{learning\_rate}        & 0.01--0.30       & 0.01--0.30       & 0.03--0.20       \\
\texttt{max\_depth}            & 3--10            & 3--10            & 4--10            \\
\texttt{n\_estimators}         & 1000             & 5000             & 5000             \\
\texttt{subsample}             & 0.50--1.00       & \textemdash      & 0.70--1.00       \\
\texttt{colsample\_bytree}     & 0.30--1.00       & \textemdash      & 0.70--1.00       \\
\texttt{gamma}                 & 0--1             & \textemdash      & \textemdash      \\
\texttt{reg\_alpha}            & 0--5             & \textemdash      & 0.0--5.0         \\
\texttt{reg\_lambda}           & 0--5             & \textemdash      & 0.0--5.0         \\
\texttt{early\_stopping\_rounds} & 20              & 20               & 20               \\
\texttt{l2\_leaf\_reg}         & \textemdash      & 1--10            & \textemdash      \\
\texttt{bagging\_temperature}  & \textemdash      & 0--1             & \textemdash      \\
\texttt{random\_strength}      & \textemdash      & 0--1             & \textemdash      \\
\texttt{num\_leaves}           & \textemdash      & \textemdash      & 15--64           \\
\texttt{min\_child\_samples}   & \textemdash      & \textemdash      & 5--30            \\
\bottomrule
\end{tabular}
\end{table}
\FloatBarrier

\begin{figure}[htbp]
    \centering
    \includegraphics[width=0.5\textwidth]{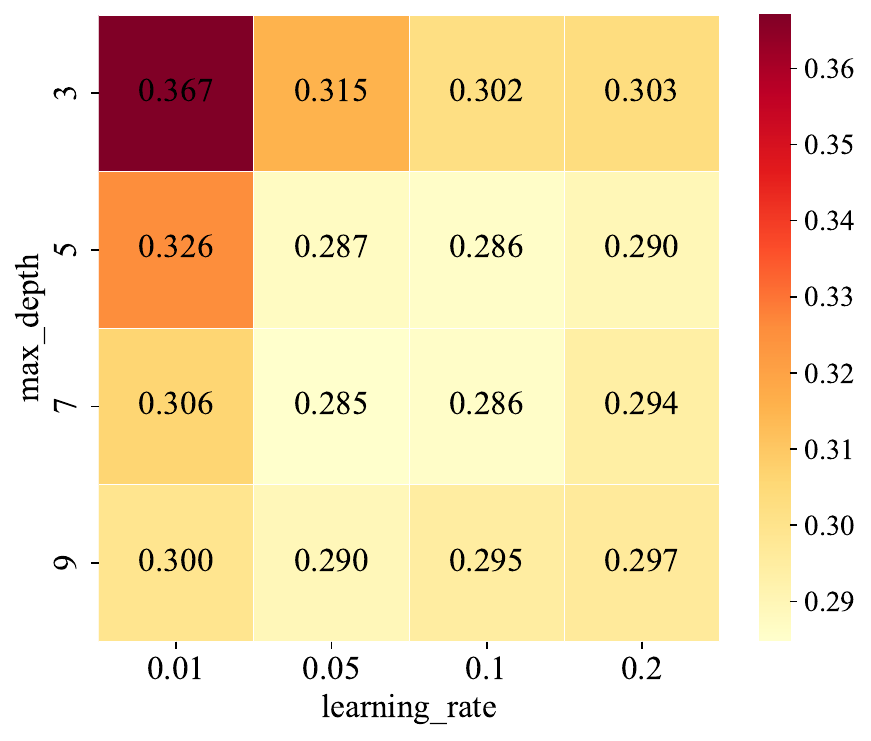}
    \caption{Mean MAE over 5-fold cross-validation for different combinations of \texttt{max\_depth} and \texttt{learning\_rate} in the XGBoost model trained on the Link Stabilizer product group.}
    \label{fig:xgb_heatmap}
\end{figure}

To examine the relationship between model performance and hyperparameter choices, we conducted a grid search over \texttt{max\_depth} and \texttt{learning\_rate} for each product group. The heatmap for Link Stabilizer for the XGBoost model is shown in Figure~\ref{fig:xgb_heatmap}, which reports mean absolute error (MAE) across 5-fold cross-validation for each combination. The best performance was achieved with \texttt{max\_depth}~$=7$ and \texttt{learning\_rate}~$=0.05$, yielding an MAE of 0.285. Overall, performance improved with increasing tree depth up to a point, particularly between depths 3 and 7, while excessively deep trees showed signs of overfitting or diminishing returns. Learning rates of 0.05 and 0.10 generally yielded better performance than both the lower value 0.01 and the higher value 0.20, indicating that intermediate learning rates strike a balance between convergence speed and model stability. As evident in the heatmap, prediction accuracy tends to improve when moving from the outer corners toward the center, particularly around the combination of moderately deep trees and mid-range learning rates.

\FloatBarrier
\FloatBarrier

\section{Results and Discussion}

This section presents a detailed analysis of the models' performance on the test sets, feature importance and practical implications of our machine learning framework for predicting manufacturing costs from 2D engineering drawings. We evaluate three gradient boosting algorithms: XGBoost, CatBoost and LightGBM on a diverse set of 24 product groups. Each model is trained using the same geometric and dimensional features derived from parsed technical drawings. Furthermore, we explore which geometric and dimensional features most strongly influence cost predictions.

Table~\ref{tab:model-performance} summarizes the performance of XGBoost, CatBoost and LightGBM models for 24 product categories using two metrics: Mean Absolute Error (MAE) and Mean Absolute Percentage Error (MAPE). Among these, MAPE is the primary evaluation criterion due to its easier interpretability in cost estimation \citep{de_Myttenaere_2016}. Considering all 24 product types,  the MAPE values range between 3.91\% and 18.51\%, with 10 groups achieving MAPE values below 10\%. It can be seen that these 10 groups do not consistently align with the largest datasets, as it is known that data size is not the only factor influencing performance \citep{hastie2009elements}. While some contain fewer than 300 samples, others include more than 1000. This suggests that model performance is not correlated only with the count of drawings in a product group. Instead, factors such as geometric feature consistency, noise levels and the inherent variability of manufacturing costs among similar parts may play a more influential role. For instance, highly standardized components with consistent geometric patterns may yield better predictive accuracy even with fewer data points, while more diverse or irregular components may require larger datasets to achieve comparable performance. 

In terms of best performance per group, XGBoost achieved the lowest MAPE in 11 product categories, followed by CatBoost in 9 and LightGBM in 4. Among the six largest product groups, which contain more than 1000 samples each, CatBoost and XGBoost each achieved the best performance in three groups. This suggests that, in this study, they outperform LightGBM when trained on larger datasets. In terms of performance margins, different models showed a clear advantage of at least 1\% in MAPE over the next best model in their respective categories, including XGBoost for Tie Rod Assembly – Tie Rod Assembly, Inner Tie Rod – Female Inner Tie Rod, Wishbones – Wishbones, and Control Arms – Forged Control Arm with Conical Hole; CatBoost for Components – Control Arm Housing with Ball Joint and Components – Machined Male Tie Rod Ends; and LightGBM for Control Arms – Forged Control Arm.

These results indicate that while all three gradient boosting models are strong contenders, XGBoost demonstrates more consistent performance, especially in groups where distinct geometric patterns or moderate sample sizes are present. On the other hand, CatBoost often excels in more complex or noisy product groups, likely due to its inherent handling of categorical features and overfitting mitigation techniques. LightGBM, though slightly behind overall, still achieves leading performance in some cases, particularly where deep trees with fewer leaves better capture geometric subtleties. Overall, the results confirm the suitability of gradient boosting algorithms for modeling the complex and nonlinear relationship between extracted geometric features and manufacturing cost in the automotive domain.

\FloatBarrier
\begin{table}[htbp]
\centering
\caption{Mean absolute error (MAE) and mean absolute percentage error (MAPE) on the test sets for each product type and model. "Size" indicates the total number of drawings in each product group; 15\% of these were used as the test set to compute the MAE and MAPE values reported here. Boldface denotes the lowest MAPE in each row.}
\label{tab:model-performance}
\begin{tabular}{@{}l c cc cc cc@{}}
\toprule
\textbf{Product group} & \textbf{Size} 
  & \multicolumn{2}{c}{\textbf{XGBoost}} 
  & \multicolumn{2}{c}{\textbf{CatBoost}} 
  & \multicolumn{2}{c}{\textbf{LightGBM}} \\ 
\cmidrule(lr){3-4} \cmidrule(lr){5-6} \cmidrule(lr){7-8}
 &  & \textbf{MAE} & \textbf{MAPE} & \textbf{MAE} & \textbf{MAPE} & \textbf{MAE} & \textbf{MAPE} \\
\midrule
Components – Machined Ball Joints with Bolt        &  591 & 0.28 & 14.04 & 0.25 & \textbf{12.26} & 0.26 & 13.00 \\
Components – Machined Female Tie Rod Ends          & 1153 & 0.13 & \textbf{7.06}  & 0.13 & 7.19           & 0.13 & 7.07  \\
Components – Machined Male Tie Rod Ends            &  329 & 0.21 & 11.56 & 0.18 & \textbf{9.89}  & 0.21 & 11.11 \\
Components – Machined Arms with Ball Joint         &  434 & 1.30 & 14.60 & 1.28 & 14.44          & 1.43 & \textbf{14.09} \\
Components – Machined Tie Rod Drag Links           &  870 & 0.10 & 9.18  & 0.10 & 9.20           & 0.10 & \textbf{9.14}  \\
Components – Machined Housing for Link Stabilizer & 1707 & 0.10 & 13.62 & 0.09 & \textbf{12.02} & 0.09 & 12.91 \\
Components – Tie Rod End Housing                   & 1146 & 0.16 & 12.92 & 0.16 & \textbf{12.50} & 0.17 & 13.17 \\
Components – Control Arm Housing with Ball Joint   &  483 & 1.07 & 18.51 & 0.88 & \textbf{13.91} & 0.91 & 15.15 \\
Tie Rod Assembly – Tie Rod Assembly                &  235 & 0.41 & \textbf{5.27}  & 0.50 & 6.32           & 0.42 & 5.35  \\
Tie Rod End – Female Tie Rod Ends                  & 1157 & 0.17 & \textbf{5.01}  & 0.18 & 5.30           & 0.18 & 5.24  \\
Ball Joint – Ball Joints with Bolt                 &  442 & 0.52 & 14.57 & 0.43 & \textbf{12.10} & 0.47 & 12.92 \\
Ball Joint – Round Ball Joints                     &  369 & 0.55 & 17.10 & 0.50 & \textbf{15.30} & 0.50 & 15.62 \\
Inner Tie Rod – Female Inner Tie Rod               &   58 & 0.52 & \textbf{13.28} & 0.72 & 17.84          & 0.67 & 17.82 \\
Inner Tie Rod – Male Inner Tie Rod                 &  902 & 0.33 & \textbf{11.55} & 0.34 & 11.76          & 0.36 & 12.33 \\
Wishbones – Wishbones                              & 1151 & 1.70 & \textbf{11.17} & 1.86 & 12.50          & 1.77 & 12.20 \\
Wishbones – Wishbones with Conical Hole            &   91 & 1.02 & \textbf{3.91}  & 1.00 & 4.00           & 1.23 & 5.72  \\
Wishbones – Wishbones without Ball Joint           &  362 & 1.34 & 10.81 & 1.29 & \textbf{10.68} & 1.39 & 11.28 \\
Control Arms – Stamped Lateral Arm                 &   64 & 1.07 & \textbf{12.66} & 1.38 & 16.17          & 1.15 & 13.14 \\
Control Arms – Aluminum Arm with Ball Joint        &  270 & 1.42 & \textbf{9.47}  & 1.51 & 10.01          & 1.71 & 11.68 \\
Control Arms – Forged Arm with Ball Joint          &  138 & 2.27 & 12.52 & 2.34 & 13.07          & 2.08 & \textbf{11.78} \\
Control Arms – Forged Control Arm                  &  108 & 2.15 & 12.30 & 1.97 & 11.51          & 1.83 & \textbf{9.74}  \\
Control Arms – Forged Control Arm with Conical Hole &   34 & 0.75 & \textbf{3.93}  & 0.82 & 5.58           & 0.89 & 4.99  \\
Control Arms – Forged Control Arm without Ball Joint &   37 & 1.23 & \textbf{4.73}  & 1.97 & 6.86           & 1.47 & 5.39  \\
Link Stabilizer                                     & 1553 & 0.29 & 10.79 & 0.28 & \textbf{10.12} & 0.30 & 11.37 \\
\bottomrule
\end{tabular}
\end{table}
\FloatBarrier

\begin{figure}[htbp]
    \centering
    \includegraphics[width=0.5\textwidth]{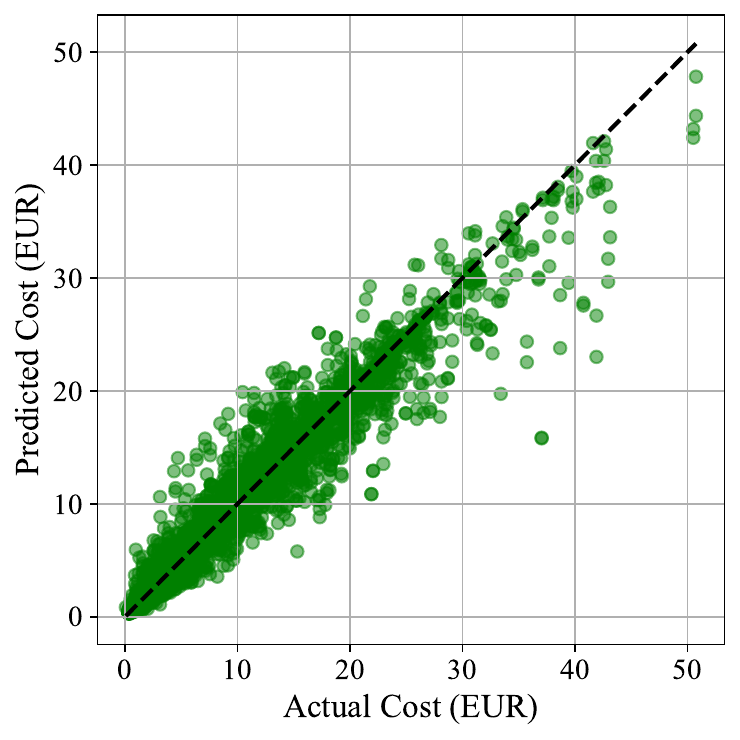}
    \caption{Actual vs. predicted unit costs across all product groups.}
    \label{fig:actual-vs-predicted}
\end{figure}

To assess overall predictive performance across all product groups, we plotted actual unit costs against predicted unit costs from the XGBoost model. Figure~\ref{fig:actual-vs-predicted} shows each engineering drawing as a point in the scatter plot. The black dashed line represents the ideal scenario where predicted and actual costs are equal. The predictions follow the diagonal, indicating strong model performance on thousands of drawings, with some outliers. These outliers are more pronounced for components with higher actual costs, where the model starts underestimating. This behavior may stem from the greater variability and complexity of expensive parts or from their limited representation in the training data.

\FloatBarrier
\FloatBarrier

\begin{figure}[htbp]
    \centering
    \includegraphics[width=\textwidth]{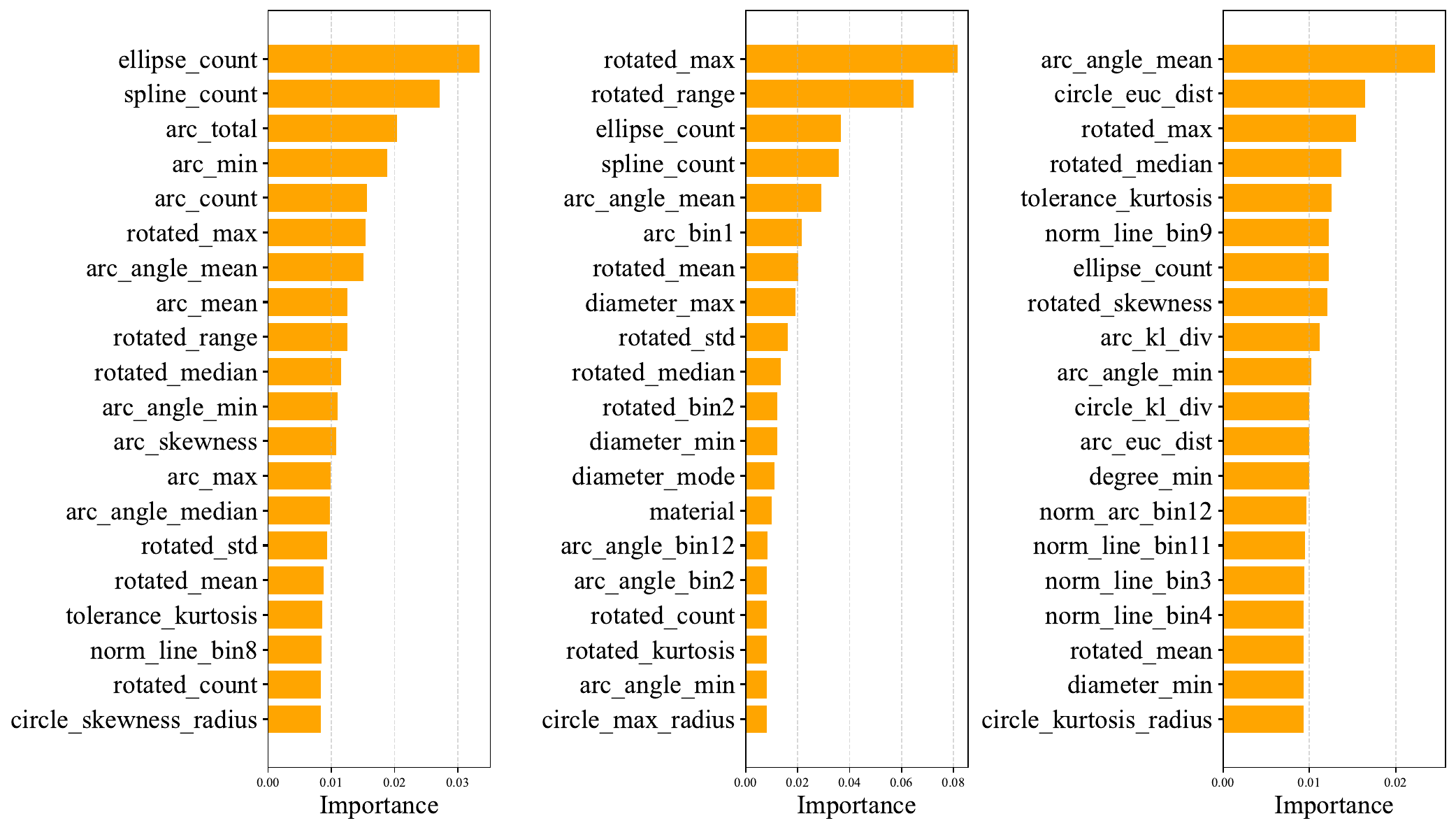}
    \caption{Top 20 most important features for each gradient boosting algorithm: XGBoost, CatBoost, and LightGBM, respectively.}
    \label{fig:feature-importance-all}
\end{figure}

We analyzed the feature importances from all 24 trained models per algorithm. For each gradient boosting model, we calculated the average normalized feature importance values based on split count for all product-specific regressors. The resulting top 20 features for each model are visualized in Figure~\ref{fig:feature-importance-all}. Considering all models, certain features consistently appeared among the top contributors. The feature \texttt{rotated\_max}, which represents the maximum measured length from rotated dimensions, was ranked highest in CatBoost with a score of 0.08 and also appeared among the top five in LightGBM and XGBoost. Features related to arc geometry such as \texttt{arc\_angle\_mean}, \texttt{arc\_angle\_min}, \texttt{arc\_mean}, and \texttt{arc\_total} were frequently ranked high in all three models, indicating that curvature information is a strong cost driver in many components. Similarly, histogram-based features such as \texttt{norm\_line\_bin8}, \texttt{norm\_line\_bin9}, and \texttt{arc\_angle\_bin12} appeared in multiple models, suggesting the relevance of distributional shape information.

\FloatBarrier
\FloatBarrier

XGBoost and CatBoost placed relatively higher emphasis on dimensional summary statistics (e.g., \texttt{rotated\_std}, \texttt{rotated\_range}, \texttt{diameter\_max}), while LightGBM also highlighted divergence-based metrics such as \texttt{arc\_kl\_div} and \texttt{circle\_kl\_div}, and distance-based metrics such as \texttt{circle\_euc\_dist} and \texttt{arc\_euc\_dist}. The inclusion of these KL divergence and Euclidean metrics implies that deviations from typical shape distributions can signal cost-relevant complexity.

Collectively, the importance rankings affirm that geometric properties, especially those related to arcs and dimensions, are critical factors influencing manufacturing cost. The recurrence of features such as \texttt{ellipse\_count}, \texttt{spline\_count}, and \texttt{rotated\_count} further supports that both shape complexity and the presence of specific curve types contribute meaningfully to cost variation.

\begin{figure}[htbp]
    \centering
    \includegraphics[width=\textwidth]{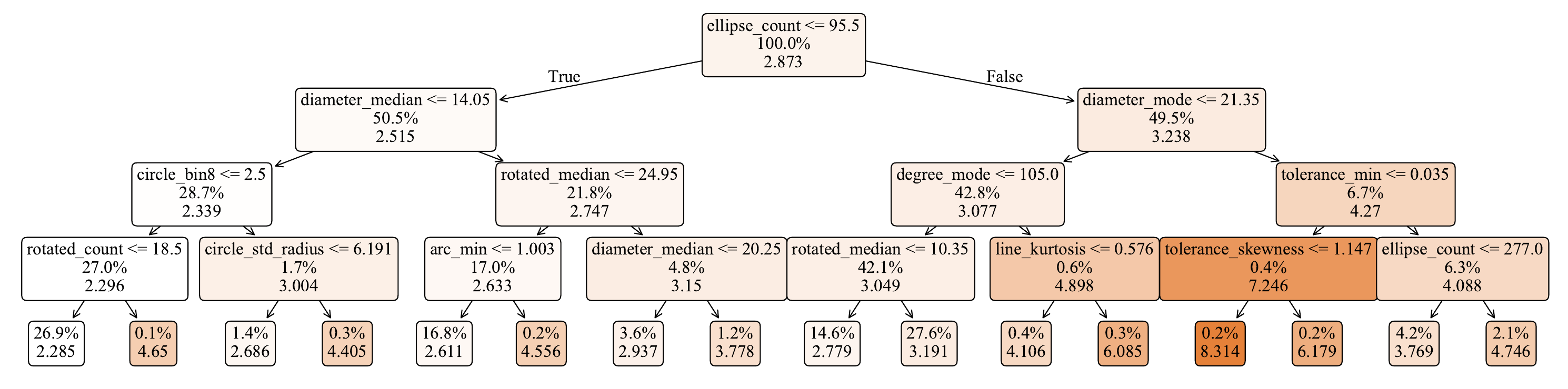}
    \caption{Simple Decision tree for the Link Stabilizer group showing feature influence on manufacturing cost. In internal nodes, the first line shows the split rule, the second line indicates the proportion of samples and the third line displays the node's prediction of cost.}
    \label{fig:decision_tree}
\end{figure}

To provide an interpretable representation of feature influence on manufacturing cost, a standalone simple decision tree regressor \citep{Breiman1984CART} was trained on the Link Stabilizer group. While gradient boosting models aggregate numerous such trees, a single representative tree provides a transparent view of the learned decision rules. Figure~\ref{fig:decision_tree} presents a simple decision tree illustrating how geometric and dimensional features interact in the cost estimation process. Each internal node displays the splitting feature (e.g., \texttt{diameter\_median}, \texttt{ellipse\_count}, \texttt{tolerance\_skewness}) and a threshold value; samples with feature values less than or equal to the threshold follow the left branch, and those above follow the right. Thresholds are expressed in scaled physical units learned during training from features extracted from 2D engineering drawings. The second line in each internal node reports the number of training samples satisfying the node condition, while the final line indicates the predicted unit manufacturing cost in Euros, corresponding to the mean cost of all samples reaching that leaf. Prominent split features such as \texttt{diameter\_median}, \texttt{ellipse\_count}, and \texttt{tolerance\_skewness} confirm earlier findings that dimensional statistics, shape counts, and distributional asymmetries are significant cost drivers. Additional splits involving \texttt{circle\_std\_radius} and \texttt{arc\_min} underscore the influence of curvature and spatial variation. This visualization complements the feature importance analysis by explicitly demonstrating the sequential decision logic that differentiates costs within a product group, thereby offering engineers actionable guidance for cost-aware design.
    
\FloatBarrier
\FloatBarrier

\begin{figure}[htbp]
    \centering
    \includegraphics[width=0.4\textwidth]{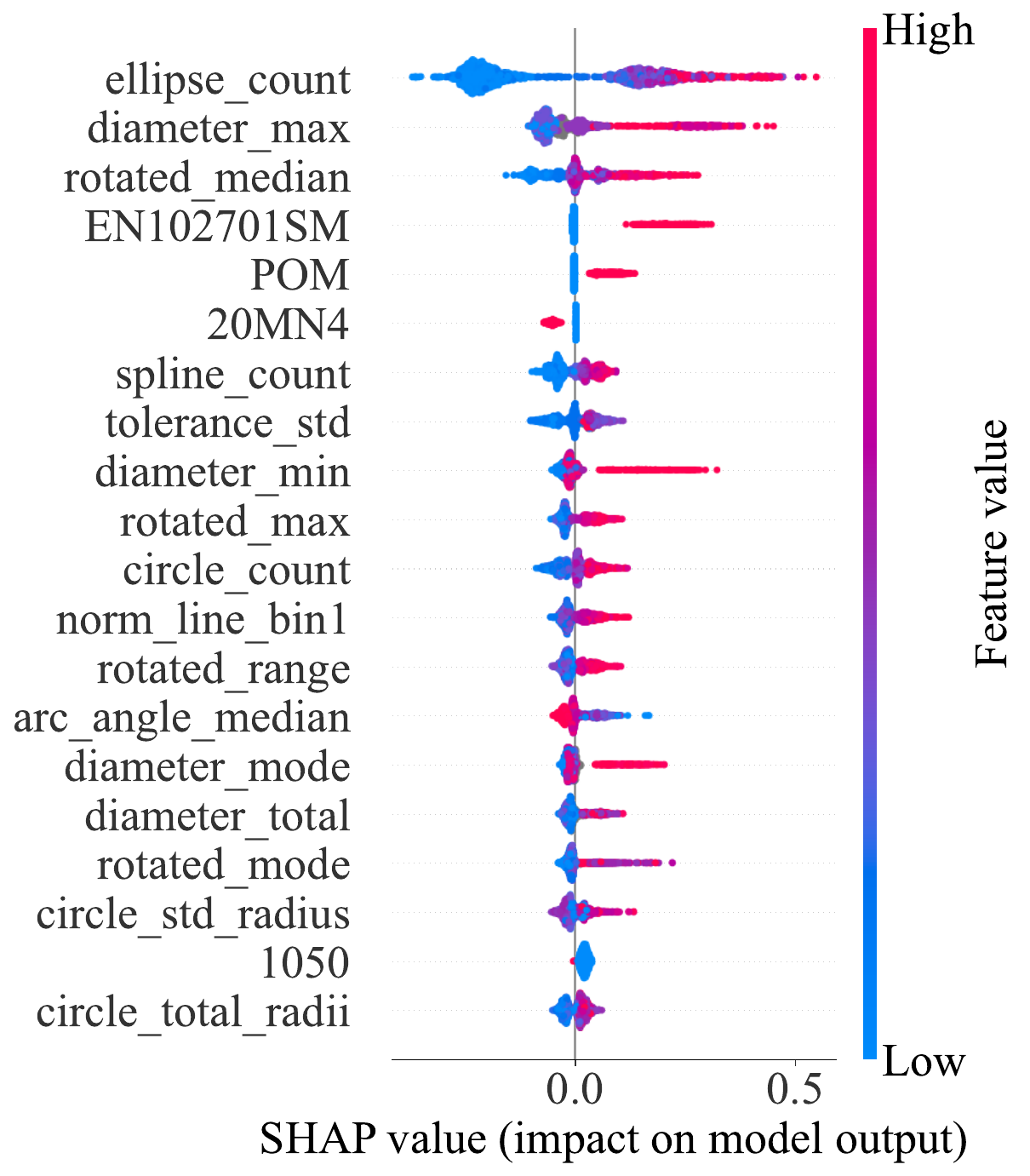}
    \caption{SHAP summary plot for the XGBoost model trained on Link Stabilizer product group.}
    \label{fig:shap_summary}
\end{figure}

We further applied Shapley Additive Explanations (SHAP) analysis to the XGBoost model trained on the Link Stabilizer group to quantify each feature's contribution to predicted cost. Figure~\ref{fig:shap_summary} displays the SHAP summary plot for the XGBoost model trained on the Link Stabilizer group. The feature \texttt{ellipse\_count} has the highest average impact on model output, suggesting that components with a greater number of ellipses tend to be associated with higher manufacturing costs. This is likely due to the added geometric complexity introduced by these shapes, which may increase tooling or machining time. \texttt{diameter\_max} and \texttt{rotated\_median} also rank high in influence, indicating that both the overall size and distribution of dimensioned features play an important role in cost estimation. Among the categorical material indicators, \texttt{TPU} appears prominently, with higher values consistently contributing to increased predicted cost, which may reflect the material's specific processing requirements. \texttt{tolerance\_std}, \texttt{diameter\_min} and other diameter-related statistics exhibit moderate but consistent impact, suggesting that local variations in dimensional precision and range influence cost sensitivity. These patterns collectively indicate that the model places significant weight on features capturing shape complexity, material selection and dimensional spread when predicting the cost of Link Stabilizer components.

\FloatBarrier
\FloatBarrier

\section{Conclusion}

This study presents a fully automated framework for predicting manufacturing costs directly from 2D DWG engineering drawings by extracting rich geometric and statistical features and applying gradient-boosted decision tree regressors. Across 24 product groups and 13,684 components, the proposed approach achieves competitive predictive accuracy, with MAPE values frequently below 10\% and without requiring explicit process planning or manual feature annotation. Comparative evaluation of XGBoost, CatBoost, and LightGBM shows that gradient boosting methods are well suited to capture nonlinear dependencies between geometric complexity, dimensional variation, material selection and manufacturing cost.

Beyond predictive performance, the integration of interpretable ML techniques—feature importance analysis, decision tree visualization and SHAP attribution—demonstrates that the most influential cost drivers align with domain knowledge. Metrics such as ellipse count, arc geometry statistics, and dimensional extrema consistently emerge as key factors, offering actionable insight for cost-aware design optimization. These explainability tools bridge the gap between data-driven estimation and engineering decision-making, enabling practitioners to identify and mitigate design features that disproportionately increase costs.

The developed machine learning framework has the potential to significantly improve cost estimation processes in real-world automotive manufacturing. By automatically extracting geometric features from 2D engineering drawings and predicting unit production costs, the model can offer early cost insights during the design or quotation phase—reducing dependency on manual expert evaluations and historical data. This approach is especially beneficial for evaluating costs in new designs or additive manufacturing contexts, where traditional cost references may be limited. Moreover, this geometry‐based, learning–driven workflow can be readily adapted to other sectors that rely on technical drawings such as civil engineering, architecture, shipbuilding or aerospace—by tailoring the feature‐extraction stage to domain‐specific entities.

The proposed system can be integrated into digital production pipelines as a decision-support tool. For example, it can be embedded into CAD platforms or ERP systems to provide instant cost feedback during the design process or to support procurement and planning decisions.

Future improvements could involve expanding the dataset through synthetic drawing generation, which could improve model robustness and address class imbalance. Additionally, incorporating spatial relationships between geometric entities (e.g., adjacency, symmetry or relative positioning) could provide richer representations and enhance performance. This could be achieved through graph-based models or by extending the current feature extraction process to capture such spatial dependencies.

\section{Acknowledgement}

This work is supported by T\"{U}B\.{I}TAK 1711 Artificial Intelligence Ecosystem Call under grant no. 3237003. Computing resources used in this work were provided by the National Center for High Performance Computing of T\"{u}rkiye (UHeM) under grant no. 1007872020.

\bibliographystyle{naturemag}
\bibliography{references}

\end{document}